\documentstyle[12pt]{article}

\newtheorem{Definition}{Definition}
\newtheorem{Corollary}{Corollary}
\newtheorem{Theorem}{Theorem}
\newtheorem{Question}{Question}

\def\PFIN{{\rm PFIN}}
\def\FIN{{\rm FIN}}
\def\EX{{\rm EX}}
\def\TXTEX{{\rm TxtEx}}
\def\bbbn{{\rm I\!N}}

\begin{document}
\bibliographystyle{plain}

\title{\bf Probabilistic Inductive Inference: 
a Survey\thanks{Parts of this work were
done at University of Latvia, supported by Latvia Science Council 
Grants 93.599 and 96.0282}}

\author{Andris Ambainis\\
	Computer Science Division\\
	University of California\\
	Berkeley, CA 94720-1776, U.S.A.\\
	e-mail:~ambainis@cs.berkeley.edu}
\date{}

\maketitle

\begin{abstract}
Inductive inference is a recursion-theoretic theory of learning,
first developed by E. M. Gold (1967).
This paper surveys developments in probabilistic inductive inference.
We mainly focus on finite inference of recursive functions, 
since this simple paradigm
has produced the most interesting (and most complex) results.
\end{abstract}

\section{Introduction}

Understanding the process of learning has always fascinated scientists.
There are several computational theories of learning.
One of the oldest theories is inductive inference established by 
Gold\cite{Gol:j:67}. This theory considers the process of 
learning from a viewpoint of the computability theory.
Unlike other theories of learning (for example,
PAC-learning\cite{Val:j:84:learnable,Kea-Vaz:b:94}), inductive inference
does not make probabilistic assumptions about the world.
However, probabilistic algorithms appear in inductive inference
and the study of probabilistic inductive inference creates
a lot of interesting problems with elements of both computability
theory and combinatorics. In this paper, we survey some of these
problems.  

We start with a general introduction to inductive inference.
Learning can be considered as a process of gathering information
about an unknown object, processing this information and 
obtaining a description of the unknown object.
Ideally, we would like to obtain a complete description of the object.

In the theory of inductive inference, objects are arbitrary recursive
(computable) functions (or recursively enumerable languages).
The reason is that any algorithmic behavior can be 
represented as a recursive (computable) function and, hence,
we obtain a model that includes any learning situation.
Throughout this paper, we only consider learning of total
recursive functions (except section \ref{lang} where 
we consider recursively enumerable languages).

The natural data about a function $f$ are its values $f(0)$, 
$f(1)$, $f(2)$, $\ldots$ and the natural representation of
these data is the sequence $\langle 0, f(0)\rangle$,
$\langle 1, f(1)\rangle$, $\langle 2, f(2)\rangle$, $\ldots$.
The most general type of description for a computable function
is a program in a universal programming language.
Any other description can be converted to this form.

This gives us the following learning model.
A learning algorithm receives the values of an unknown 
function $f$ in the natural order: $\langle 0, f(0)\rangle$,
$\langle 1, f(1)\rangle$, $\langle 2, f(2)\rangle$, $\ldots$
and produces a program $h$. The algorithm succeeds on $f$
if the program $h$ computes $f$.

We will compare the classes of functions identifiable by
probabilistic algorithms with different probabilities of 
correct answer.

\section{Definitions}

Next, we introduce the formal notation and definitions used in
this paper. For more background information, see \cite{Rog:b:87} for
recursive function (computability) theory, \cite{Sie:b:65,Kur-Mos:b:67}
for set theory and \cite{Ang-Smi:j:83,Osh-Sto-Wei:b:86:stl} for
inductive inference.

A {\em learning machine} is an algorithmic device that reads
values of a function $f$: $f(0)$, $f(1)$, $\ldots$.
Having seen finitely many values of the function
it can output a conjecture.
{\em A conjecture} is a program in some fixed acceptable 
programming system\cite{Mac-You:b:78,Rog:b:87}.

It makes sense to allow the learning machine to revise its conjecture.
In this case, the last conjecture output by the algorithm should be correct
but intermediate conjectures may be wrong. This increases the power of
the learning algorithm. Also, this can be motivated by the fact
that humans learning a complex behavior (for example, foreign language 
or driving), do not learn it completely at once.
Rather, they first learn a part of it, then extend it by
learning more. This model where an unlimited number of conjectures
is allowed is called {\em learning in the limit}\cite{Gol:j:67}.

\begin{Definition}
\cite{Gol:j:67}
\begin{enumerate}
\item[(a)]
A deterministic learning machine $M$ $\EX$-identifies 
(identifies in the limit) a function $f$
if, given $f(0), f(1)$, $\ldots$ it outputs a sequence of programs
$h_0, h_1, \ldots$ such that, for some $i$, $h_i=h_{i+1}=h_{i+2}=\ldots$
and $h_i$ is a program computing $f$.
\item[(b)]
$M$ $\EX$-identifies a set of functions $U$ 
if it $\EX$-identifies all $f\in U$.
\item[(c)]
$EX$ denotes the set of all sets $U$ that are $\EX$-identifiable.
\end{enumerate}
\end{Definition}

A model where only one conjecture is allowed and it must be correct
has been studied as well. This model is called finite learning.
It is more restricted and (in most contexts) simpler.

\begin{Definition}
\cite{Gol:j:67}
\begin{itemize}
\item[(a)]
A deterministic learning machine $M$ 
{\em finitely identifies ($\FIN$-identifies)} 
a function $f$ if, receiving $f(0)$, $f(1)$, $\ldots$ as the input, 
it produces a program computing function $f$.
\item[(b)]
$M$ $\FIN$-identifies a set of functions $U$ if
it $\FIN$-identifies any function $f\in U$.
\item[(c)]
A set of functions $U$ is called {\em $\FIN$-identifiable}
if there exists a deterministic learning machine that identifies $U$.
The collection of all $\FIN$-identifiable sets is denoted by $\FIN$.
\end{itemize}
\end{Definition}

The problems that we consider are fairly simple for probabilistic
learning in the limit (cf. \cite{Pit:j:89,Pit-Smi:j:88,Smi:j:82})
but are much more complicated (and more interesting) for finite 
learning. Therefore, in this survey, we focus on finite learning.
References to work on other models of inductive inference
are given in sections \ref{lang} and \ref{Related}.
Next, we define identification by probabilistic machines.
We define it for FIN but the definition carries over to
EX and other paradigms as well.

\begin{Definition}
\begin{itemize}
\item[(a)]
A probabilistic learning machine $M$ $\langle p\rangle\FIN$-identifies
($\FIN$-identifies with probability $p$) the set of functions $U$ if, 
for any function $f\in U$ the probability that $M$ $FIN$-identifies
$f$ is at least $p$.
\item[(b)]
The collection of all $\langle p \rangle\FIN$-identifiable
sets is denoted by $\langle p \rangle\FIN$.
\end{itemize}
\end{Definition}

{\em Team identification} is another idea closely related to probabilistic
identification. A team is just a finite set of learning machines
$\{M_1, M_2, \ldots, M_s\}$. 

\begin{Definition}
\begin{itemize}
\item[(a)]
A team $M$ $[r, s]\FIN$-identifies the function $f$
if at least $r$ of learning machines
$M_1$, $\ldots$, $M_s$ $\FIN$-identify $f$.
\item[(b)]
The collection of all $[r, s]\FIN$-identifiable
sets is denoted by $[r, s]\FIN$.
\end{itemize}
\end{Definition}

It is easy to see that $[r, s]\FIN\subseteq \langle \frac{r}{s}\rangle\FIN$.
(Just choose one of the machines in the team uniformly at random and
simulate.) In some cases, the opposite is also true and every probabilistic
machine can be simulated by a team. 

The main goal of research in probabilistic inductive inference
is determining how $\langle p\rangle\FIN$ depends on the accepting
probability $p$. Formally, it means describing the {\em probability hierarchy}.

\begin{Definition}
The probability hierarchy for $\FIN$ is the set
of all points $p$ such that there is $U\in\langle p\rangle\FIN$
but $U\notin\langle p+\epsilon\rangle\FIN$ for $\epsilon>0$.
\end{Definition}

\section{Explicit results for FIN}
\label{Third}

Probabilistic $\FIN$-identification was first studied by 
Freivalds\cite{Fre:c:79:prob-fin}. He showed that any probabilistic learning
machine with the probability of correct answer above $2/3$ can be
replaced by an equivalent deterministic machine. He also
characterized the probability hierarchy for $\FIN$
between $1/2$ and $2/3$.

\begin{Theorem}
\cite{Fre:c:79:prob-fin}
\label{Above12}
\begin{enumerate}
\item[(a)]
If $p>2/3$, then $\langle p\rangle\FIN =\FIN$.
\item[(b)]
$\langle 2/3\rangle FIN\neq FIN$.
\item[(c)]
If $n/(2n-1)\geq p>(n+1)/(2n+1)$, then 
$\langle p\rangle\FIN=\langle n/(2n-1)\rangle\FIN=[n, 2n-1]\FIN$.
\item[(d)]
$\langle (n+1)/(2n+1)\rangle\FIN\neq \langle n/(2n-1)\rangle\FIN$.
\end{enumerate}
\end{Theorem}

It also makes sense to consider probabilistic algorithms with
the probability of correct answer 1/2 and below
because there are infinitely many outputs and, hence, 
even designing an algorithm that gives the correct answer with probability
$\epsilon$ (for an arbitrarily small fixed $\epsilon>0$) may be nontrivial.
Here, the first result was a surprising discovery that
a ``2 out of 4" team is more powerful than a ``1 out of 2" team.

\begin{Theorem}
\cite{Jai-Sha-Vel:j:95:teammind}
\label{At12}
\begin{enumerate}
\item[(a)]
There is a set of functions $U$ such that $U\in[2, 4]\FIN$ but
$U\notin[1, 2]\FIN$.
\item[(b)]
$[1, 2]\FIN=[3, 6]\FIN=[5, 10]\FIN=\ldots$ and 
$[2, 4]\FIN=[4, 8]\FIN=[6, 12]\FIN=\ldots$.
\item[(c)]
$\langle 1/2\rangle\FIN=[2,4]\FIN$.
\end{enumerate}
\end{Theorem}

We see that the power of a team depends not only
on the ratio of machines that must succeed but also on the number
of machines in the team. 

The next step was moving below probability 1/2.

\begin{Theorem}
\label{Below12}
\cite{Dal-Kal-Vel:j:95}
Let $p_0=\frac{1}{2}$, $p_1=\frac{24}{49}$, $p_2=\frac{20}{41}$,
$p_3=\frac{18}{37}$, $p_4=\frac{17}{35}$.
Then, for all $i\in\{0, 1, 2, 3\}$
\begin{enumerate}
\item[(a)]
For all $x\in]p_{i+1}, p_i]$, $\langle x\rangle\FIN=\langle p_i\rangle\FIN$, and
\item[(b)]
$\langle p_i\rangle\FIN\neq\langle p_{i+1}\rangle\FIN$.
\end{enumerate}
\end{Theorem}

Each of these cutpoints was proven separately and
there seemed to be no formula or unifying proof argument connecting them. 
It took several years to obtain a more general result.

\begin{Theorem}
\label{Below12a}
\cite{Dal-Kal:c:97}
The probability hierarchy for $\FIN$ in the interval 
$[\frac{12}{25}, \frac{1}{2}]$ is 
$\{\frac{12m-16}{25m-34} | m\geq 2\}\cup
\{\frac{24}{49}, \frac{20}{41}, \frac{17}{35}, \frac{15}{31},
\frac{27}{56}\}$.
\end{Theorem}

Thus, it appeared that there was a formula ($\frac{12m-16}{25m-34}$)
and a general argument for this interval. It only was obscured
by exceptions from this formula at the beginning.
With probabilities getting smaller, progress became more and more
difficult. The full proof of Theorem \ref{Below12a} was more than 100 pages 
long. On the other hand, it only described the situation for
the interval $[\frac{12}{25},\frac{1}{2}]$.

\section{Explicit results for PFIN}

One of the approaches to this situation was considering
Popperian FINite identification(PFIN), a restricted version of
FIN. FIN allows two types of errors
on functions that are not identified by a machine.
These are
\begin{enumerate}
\item
Errors of commission. The program output by a machine $M$ 
produces a value different from the value of the input function.
\item
Errors of omission. The program output by $M$ does not
halt on some input.
\end{enumerate}
Errors of omission are ones that cause most trouble.
The reason is that, given a program $h$
output by a machine $M$, we cannot tell whether $h$ halts on input $x$.
If we eliminate them, the model becomes simpler and still
remains interesting.

\begin{Definition}
\cite{Min:j:76}
A learning machine $M$ is Popperian if it does not make
errors of omission (i.e., if all conjectures on all inputs
are programs computing total functions).
\end{Definition}

\begin{Definition}
\begin{enumerate}
\item[(a)]
A set of functions $U$ is PFIN-identifiable if there is
a Popperian machine $M$ that $\FIN$-identifies $U$.
\item[(b)]
PFIN denotes the collection of all PFIN-identifiable sets.
\end{enumerate}
\end{Definition}

Probabilistic and team PFIN-identification are introduced 
similarly. It is important that the requirement about
learners outputting only programs computing total 
recursive functions is absolute, i.e.
\begin{enumerate}
\item
All conjectures of all machines in a $\PFIN$-team
must be programs computing total recursive functions.
\item
A probabilistic $\PFIN$-machine is not allowed to output 
a program which does not compute a total recursive function
even with a very small probability.
\end{enumerate}

Daley, Kalyanasundaram and Velauthapillai
\cite{Dal-Kal-Vel:c:92:popper,Dal-Kal:c:93:popper} proved counterparts of
Theorems \ref{Above12}, \ref{At12} and \ref{Below12} for $\PFIN$.
The situation for probabilities greater than or equal to 1/2
was precisely the same as for $\FIN$, only the proofs became simpler.
For probabilities smaller than 1/2, two sequences of points
where the power of probabilistic machines changed were discovered. 
The first, $\frac{4n}{9n-2}$, started at $\frac{12}{25}$ and 
converged to $\frac{4}{9}$.

\begin{Theorem}
\label{PFIN1}
\cite{Dal-Kal-Vel:c:92:popper}
\begin{enumerate}
\item[(a)]
The probability hierarchy for PFIN in the interval $[\frac{1}{2}, 1]$ is
$\{\frac{n}{2n-1}|n\geq 1\}$.
\item[(b)]
The probability hierarchy for PFIN in the interval $[\frac{4}{9}, \frac{1}{2}]$ 
is $\{\frac{4n}{9n-2}|n\geq 2\}$.
\end{enumerate}
\end{Theorem}

The second sequence was more complicated. It was actually a union of three 
simpler sequences corresponding to three different ways how machines
in a team can behave.

\begin{Theorem}
\label{PFIN2}
\cite{Dal-Kal:c:93:popper}
The probability hierarchy for PFIN in the interval $[\frac{3}{7}, \frac{4}{9}]$
is $\{\frac{6n}{14n-3}|n\geq 6\}\cup\{\frac{3n}{7n-1}|n\geq 12\}
\cup\{\frac{8n}{19n-4}|4\leq n\leq 11\}$.
\end{Theorem}

However, even for Popperian learning, things were getting more complicated
as the probabilities decreased (this can be observed both by just comparing
the sequences of probabilities in Theorems \ref{PFIN1} and \ref{PFIN2}
and by looking at the arguments that were used to prove these theorems).
As result of that, the authors of \cite{Dal-Kal:c:93:popper} wrote that 
the prospects of determining all cutpoints are bleak even for 
the interval $[2/5, 1/2]$.

\section{General results for PFIN}
\label{PGen}

An alternative approach was proposed in \cite{Amb:c:96}.
Instead of trying to find all cutpoints explicitly,
\cite{Amb:c:96} focused on studying the general properties
of the whole probability structure. 

The first step was describing existing diagonalization constructions
(i.e. constructions proving that there is $U\in \langle p\rangle\PFIN$
such that $U\notin\langle p+\epsilon\rangle\PFIN$ for $\epsilon>0$)
in a general form.

\begin{Theorem}
\cite{Amb:c:96,Kum:c:94}
Let $P_{\PFIN}$ be the probability hierarchy for $\PFIN$
and $p_1, \ldots, p_s\in P_{PFIN}$. Let $p\in[0, 1]$.
If there are $q_1\geq 0, \ldots, q_s\geq 0$ such that
\begin{enumerate}
\item
$q_1+q_2+\ldots+q_s=p$;
\item
$\frac{p}{q_i+1-p}=p_i$ for $i=1, \ldots, s$,
\end{enumerate}
then $p\in P_{PFIN}$.
\end{Theorem}

This led to a conjecture that $P_{PFIN}$ is equal to the set $A$ defined
as follows.
\begin{enumerate}
\item
$1\in A$
\item
\label{Rule}
If $p_1, p_2, \ldots, p_s \in A$ and $p\in[0, 1]$ is such that
there exist $q_1, \ldots, q_s\in[0,1]$ satisfying
\begin{enumerate}
\item
$q_1+q_2+\ldots+q_s=p$;
\item
$\frac{p}{q_i+1-p}=p_i$ for $i=1, \ldots, s$,
\end{enumerate}
then $p\in A$.
\end{enumerate}

Indeed, $A=P_{PFIN}$ and the first step in proving that was observing some
structural properties of this set.

\begin{Definition}
\cite{Sie:b:65,Kur-Mos:b:67}
A set $A$ is well-ordered if there is no infinite strictly
increasing sequence of elements of $A$.
A set $A$ is well-ordered in decreasing order
if there is no infinite strictly
increasing sequence of elements of $A$.
\end{Definition}

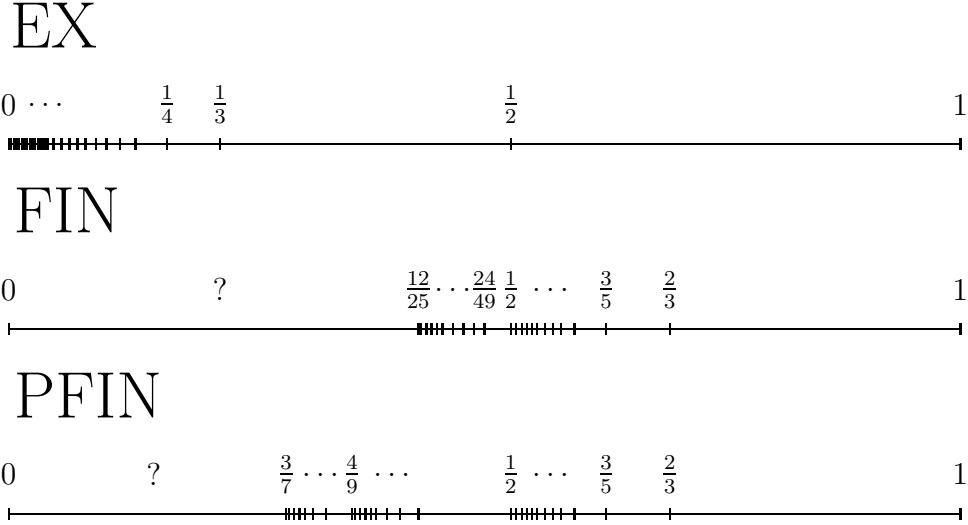
\begin{figure*}[tb]
\label{Figure}
\begin{center}
\begin{picture}(380,250)
\multiput(20,20)(0,70){3}{\line(1,0){360}}
\multiput(20,18)(0,70){3}{\line(0,1){4}}
\multiput(380,18)(0,70){3}{\line(0,1){4}}
\multiput(20,35)(0,70){3}{\makebox(0,0){$0$}}
\multiput(380,35)(0,70){3}{\makebox(0,0){$1$}}
\multiput(20,18)(0,70){2}{\line(0,1){4}}
\multiput(270,35)(0,70){2}{\makebox(0,0){$\frac{2}{3}$}}
\multiput(270,18)(0,70){2}{\line(0,1){4}}
\multiput(246,35)(0,70){2}{\makebox(0,0){$\frac{3}{5}$}}
\multiput(246,18)(0,70){2}{\line(0,1){4}}
\multiput(234,18)(0,70){2}{\line(0,1){4}}
\multiput(229,18)(0,70){2}{\line(0,1){4}}
\multiput(226,18)(0,70){2}{\line(0,1){4}}
\multiput(223,18)(0,70){2}{\line(0,1){4}}
\multiput(220,18)(0,70){2}{\line(0,1){4}}
\multiput(218,18)(0,70){2}{\line(0,1){4}}
\multiput(216,18)(0,70){2}{\line(0,1){4}}
\multiput(214,18)(0,70){2}{\line(0,1){4}}
\multiput(212,18)(0,70){2}{\line(0,1){4}}
\multiput(210,18)(0,70){3}{\line(0,1){4}}
\multiput(210,35)(0,70){3}{\makebox(0,0){$\frac{1}{2}$}}
\multiput(225,35)(0,70){2}{\makebox(0,0){$\ldots$}}
\put(200,88){\line(0,1){4}}
\put(200,105){\makebox(0,0){$\frac{24}{49}$}}
\put(196,88){\line(0,1){4}}
\put(192,88){\line(0,1){4}}
\put(188,88){\line(0,1){4}}
\put(184,88){\line(0,1){4}}
\put(182,88){\line(0,1){4}}
\put(180,88){\line(0,1){4}}
\put(178,88){\line(0,1){4}}
\put(176,88){\line(0,1){4}}
\put(175,88){\line(0,1){4}}
\put(175,105){\makebox(0,0){$\frac{12}{25}$}}
\put(188,105){\makebox(0,0){$\ldots$}}
\put(100,105){\makebox(0,0){?}}
\put(175,18){\line(0,1){4}}
\put(168,18){\line(0,1){4}}
\put(163,18){\line(0,1){4}}
\put(159,18){\line(0,1){4}}
\put(157,18){\line(0,1){4}}
\put(155,18){\line(0,1){4}}
\put(153,18){\line(0,1){4}}
\put(151,18){\line(0,1){4}}
\put(150,18){\line(0,1){4}}
\put(150,35){\makebox(0,0){$\frac{4}{9}$}}
\put(165,35){\makebox(0,0){$\ldots$}}
\put(140,18){\line(0,1){4}}
\put(135,18){\line(0,1){4}}
\put(132,18){\line(0,1){4}}
\put(130,18){\line(0,1){4}}
\put(128,18){\line(0,1){4}}
\put(126,18){\line(0,1){4}}
\put(125,18){\line(0,1){4}}
\put(125,35){\makebox(0,0){$\frac{3}{7}$}}
\put(138,35){\makebox(0,0){$\ldots$}}
\put(75,35){\makebox(0,0){?}}
\put(100,158){\line(0,1){4}}
\put(100,175){\makebox(0,0){$\frac{1}{3}$}}
\put(80,158){\line(0,1){4}}
\put(80,175){\makebox(0,0){$\frac{1}{4}$}}
\put(68,158){\line(0,1){4}}
\put(62,158){\line(0,1){4}}
\multiput(57,158)(-4,0){3}{\line(0,1){4}}
\multiput(46,158)(-3,0){3}{\line(0,1){4}}
\multiput(37,158)(-2,0){5}{\line(0,1){4}}
\multiput(35,158)(-1,0){15}{\line(0,1){4}}
\put(35,175){\makebox(0,0){\ldots}}
\put(50,65){\makebox(0,0){\huge $\PFIN$}}
\put(42,135){\makebox(0,0){\huge $\FIN$}}
\put(37,205){\makebox(0,0){\huge $\EX$}}
\end{picture}
\end{center}
\caption{The probability hierarchies for $\EX$, $\FIN$ and $\PFIN$}
\end{figure*}

Figure 1 shows the known parts of probability hierarchies for
$\EX$, $\FIN$ and $\PFIN$.
It is easy to see that  all are well-ordered in decreasing order.
The set $A$ defined above is well-ordered as well.

\begin{Theorem}
\cite{Amb:c:96}
The set $A$ is well-ordered and has a system of notations.
\end{Theorem}

A system of notations is an algorithmic description for
a well-ordered set. It allows to find preceding elements, 
given one element. This notion was introduced by Kleene
for constructive ordinals\cite{Kle:j:38} and extended
to sets of reals (like $A$) in \cite{Amb:c:96}.
Well-orderedness is crucial because it
allows to use induction over elements of the set $A$.
Having the system of notations is important to
make this induction algorithmic.
Using well-orderedness and system of notations,
\cite{Amb:c:96} showed the following result.

\begin{Theorem}
\cite{Amb:c:96}
Let $p\in A$ and $p'<p$ be such that there is no $p''\in A$
with $p'\leq p''<p$. Then, $\langle p\rangle\PFIN=\langle p'\rangle\PFIN$.
\end{Theorem}

\begin{Corollary}
\cite{Amb:c:96}
$A=P_{PFIN}$.
\end{Corollary}

This approach gives two other interesting results.

\begin{Theorem}
\cite{Amb:c:96}
The probability structure of $\PFIN$ is decidable, i.e. there is an algorithm
that receives two probabilities $p_1$ and $p_2$ and
answers whether $\langle p_1\rangle \PFIN=\langle p_2\rangle \PFIN$.
\end{Theorem}

\begin{Theorem}
\cite{Amb:c:96}
Let $p\in P_{PFIN}$. Then, there is a $k$ such that $[pk, k]\PFIN=
\langle p\rangle\PFIN$.
\end{Theorem}

Thus, teams of different size can have different learning power
(the counterpart of Theorem \ref{At12} in \cite{Dal-Kal:c:93:popper}) 
but we always have the ``best" team size 
such that team of this size can simulate any probabilistic machine
(and hence, team of any other size with the same success ratio).

Finally, it is also possible to determine the precise ordering type
of the probability hierarchy. The table below shows how the 
complexity of the ordering increases when probabilities decrease.

\medskip
\begin{center}
\begin{tabular}{|c||c|}
\hline
Interval & Ordering type of the probability hierarchy \\
\hline
\hline
$[\frac{1}{2}, 1]$ & $\omega$ \\
\hline
$[\frac{4}{9}, 1]$ & $2\omega$ \\
\hline
$[\frac{3}{7}, 1]$ & $3\omega$ \\
\hline
$[\frac{2}{5}, 1]$ & $\omega^2$ \\
\hline
$[\frac{3}{8}, 1]$ & $\omega^3$ \\
\hline
$[\frac{1}{3}, 1]$ & $\omega^{\omega}$ \\
\hline
$[\frac{1}{4}, 1]$ & $\omega^{\omega^{\omega}}$ \\
\hline
$[0, 1]$ & $\epsilon_0$ \\
\hline
\end{tabular}
\end{center}
\medskip

$\omega$ is the ordering type corresponding to a single infinite sequence
(2/3, 3/5, 4/7, $\ldots$), $k\omega$ is the ordering type of a set
consisting of $k$ infinite sequences. $\omega^2$ is the ordering type of
a set consisting of infinite sequence of sequences and $\omega^3$ is the
ordering type of an infinite sequence of $\omega^2$-type sets.
$\omega^\omega$ is the limit of $\omega$, $\omega^2$, $\omega^3$, $\ldots$.
Further ordering types can be defined similarly\cite{Sie:b:65,Kur-Mos:b:67}. 
The last one, $\epsilon_0$,
is the limit of
\[ \omega, \omega^{\omega}, \omega^{\omega^\omega}, \ldots \]
and is considered to be so big that it is hard to find any intuitive 
description for it\footnote{It is also known\cite{Sie:b:65} that $\epsilon_0$ is the ordering type
of the set of all expressions possible in first-order arithmetic but
this does not look very relevant to our inductive inference result.}. 
This shows that the explored part of $\PFIN$-hierarchy
(the interval $[\frac{3}{7}, 1]$, the ordering type $3\omega$) is very
simple compared to the entire hierarchy.
This result can be also considered as a partial explanation
why it is unrealistic to find explicit values for all points
in the probability hierarchy.

\section{General results for FIN?}
\label{FGen}

An easy corollary of results in \cite{Amb:c:96} is

\begin{Theorem}
If $\langle p_1\rangle\PFIN\neq \langle p_2\rangle\PFIN$, then 
$\langle p_1\rangle\FIN\neq \langle p_2\rangle\FIN$.
\end{Theorem}

Thus, any diagonalization argument that works for PFIN will 
work for FIN as well. This means that the probability hierarchy
for FIN is at least as complicated as for PFIN. 
(In fact, it is more complicated because there are points
(like 24/49) that are not contained in the PFIN hierarchy but
appear in the FIN hierarchy.)

There have been several attempts to move beyond explicit probabilities and
to find general proof methods for FIN. Daley and Kalyanasundaram
\cite{Dal-Kal:c:95,Dal-Kal:c:97} have developed a set of reduction
arguments (techniques to reduce the problems about inclusions for smaller 
probabilities to already solved problems for inclusion at bigger probabilities).
These arguments were essential to proving Theorem \ref{Below12a}.
They also were able to explain ``the strange probabilities"
of Theorem \ref{Below12}. Yet, the number of cases they had to handle
is huge and, for further progress, even more general techniques are 
necessary. 

Similar reduction arguments for PFIN\cite{Dal-Kal:c:93:popper} were the 
foundation for general results of \cite{Amb:c:96}. So, we may expect that 
methods of \cite{Dal-Kal:c:97} could serve as a foundation for similar results
for FIN. 

Another approach was taken by \cite{Aps:th:98,Aps-Fre-Smi:c:97,Amb-Aps-Fre:j:99} 
who defined {\em asymmetric teams}, a generalization of usual teams. 
\cite{Amb-Aps-Fre:c:97} showed that a more general result about asymmetric
teams (well-quasi-orderedness) would imply well-orderedness 
and decidability of the FIN-hierarchy. They also claimed a proof of
well-quasi-orderedness for asymmetric FIN teams. However, a bug was
discovered in this proof and it turned out that asymmetric FIN-teams
are not well-quasi-ordered\cite{Aps:th:98}. This suggests that, if
it is possible to prove a counterpart of results in section \ref{PGen}
for FIN, then the proof should use properties that are specific to
traditional teams.

\section{Other problems about FIN and PFIN}
\label{oracle}

Besides finding the cutpoints, there are other problems
about probabilistic and team learning that are worth studying.
One of them relates probabilistic and team learning to oracle
computation.

Assume that we have two teams (or probabilistic machines)
and one of them is weaker than the other. 
If we allow the weaker team (probabilistic machine)
to access some oracle (for example, $K$, the oracle for 
the halting problem), we increase the power of this team
and it may be able to learn everything that the stronger
team can learn. Kummer\cite{Kum:c:94}\footnote{Related
problems about oracles have been also studied in 
\cite{For-Gas-etal:j:94,Kum-Ste:j:96}} studied the following 
problem: given $a, b, c, d$ such that $[a, b]\FIN\not\subseteq
[c, d]\FIN$, what is the class of oracles $A$ such that
$[a, b]\FIN\subseteq [c, d]\FIN[A]$?
($[c, d]\FIN[A]$ denotes the collection of sets
of functions that are identifiable by a $[c, d]\FIN$-team
with access to oracle $A$)

We summarize his results in two theorems below.
The first theorem partitions $a, b, c, d$ into such that
$[a, b]\FIN\subseteq [c, d]\FIN[A]$ for some $A$
and such that $[a, b]\FIN\not\subseteq [c, d]\FIN[A]$
for all $A$. It also shows that, whenever $A$ exists,
the halting oracle $K$ can be used as $A$.

\begin{Theorem}
\cite{Kum:c:94}
\begin{enumerate}
\item
If there is $k\in\bbbn$ such that $\frac{m}{n}\leq\frac{1}{k}<\frac{m'}{n'}$,
then $[m, n]\FIN\not\subseteq [m', n']\FIN[A]$ for any oracle $A$.
\item
If $\frac{1}{k+1}<\frac{m}{n}$ and $\frac{m'}{n'}\leq\frac{1}{k}$
for some $k$, then $[m, n]\FIN\subseteq [m', n']\FIN[A]$ for any
$A$ such that $K\leq_{T} A$ (i.e., the halting oracle $K$ is Turing-reducible
to $A$).
\end{enumerate}
\end{Theorem}

The second theorem considers the question whether oracles weaker than
$K$ can be used as $A$ in some cases. 
For this result, we need some extra definitions.

Let $M_1, M_2, \ldots$ be an enumeration of all Turing machines
and $\varphi_i$ be the partial function computed by $M_i$.

\begin{Definition}
\label{PA}
$PA$ denotes the set of all oracles $A$ such that given $A$,
there is a function $f(x, y)$ that is computable with the oracle $A$
and:
\begin{enumerate}
\item
If $\varphi_x(y)=0$ or $\varphi_x(y)=1$, then $f(x, y)=\varphi_{x}(y)$.
\item
Otherwise, $f(x, y)$ can be anything but it must be defined
(even if $\varphi_{x}(y)$ is undefined).
\end{enumerate}
\end{Definition}

In other words, an oracle in $PA$ can be used to extend any partial
recursive function to a total recursive function that is consistent
with the original function.
Equivalently, $PA$ can be defined as the set of all oracles $A$ that 
are Turing-equivalent to a complete and consistent extension of
Peano arithmetic\cite{Kum:c:94,Joc:b:89,Odi:b:89}.
If $K$ reduces to $A$, then $A\in PA$. However, the converse is
not true\cite{Odi:b:89}.

\begin{Theorem}
\cite{Kum:c:94}
\begin{enumerate}
\item
Let $m, n$ be such that $[m, n]\PFIN\not\subseteq [m', n']\PFIN$.
Then, $[m, n]\PFIN\subseteq [m', n']\PFIN[A]$ if and only if
$[m, n]\FIN\subseteq [m', n']\FIN[A]$ if and only if 
$K\leq_{T} A$.
\item
$[24, 49]\FIN\subseteq [1, 2]\FIN[A]$ if and only if $A\in PA$.
\end{enumerate}
\end{Theorem}

Thus, we see that a weaker oracle may suffice
because there are $A\in PA$ such that $K$ is not Turing-reducible to $A$
\cite{Odi:b:89}.
Kummer\cite{Kum:c:94} asked whether these two possibilities
(we need an oracle $A$ such that $K\leq_T A$ or any $A\in PA$ suffices)
are the only ones. We have a partial answer\cite{Amb:m:98}.

\begin{Definition}
\label{PA1}
$PA'$ denotes the set of all oracles $A$ such that given $A$,
there is a function $f(x, y)$ that is computable with an oracle $A$
with the following properties:
\begin{enumerate}
\item
If $\varphi_x(y)=0$ or $\varphi_x(y)=1$, then $f(x, y)=\varphi_x(y)$.
\item
Otherwise, $f(x, y)$ can be anything but it must be defined
(even if $\varphi_x(y)$ is undefined).
\item
If, for some $x$, there is at most one $y$ such that $\varphi_x(y)=1$, 
then $f(x, y)=1$ for at most one $y$.
\end{enumerate}
\end{Definition}

\begin{Theorem}
\cite{Amb:m:98}
For any $a, b, c, d$ such that $[a, b]\FIN\not\subseteq [c, d]\FIN$,
the set of oracles $A$ such that $[a, b]\FIN\subseteq [c, d]\FIN[A]$
is one of the following:
\begin{enumerate}
\item
The empty set.
\item
The set of all $A$ such that $K\leq_T A$.
\item
$PA$ (see definition \ref{PA}).
\item
$PA'$ (see definition \ref{PA1}).
\end{enumerate}
\end{Theorem}

It is easy to see that $PA\subseteq PA'\subseteq \{A : K\leq_T A\}$.
However, we do not know whether $PA'$ coincides with $PA$,
$\{A :K\leq_T A\}$ or is different from both of them.

Other properties of the probability hierarchy deserve studying as
well. For example, \cite{Amb:c:96} asked how close are points
of the $\PFIN$-hierarchy one to another.

\begin{Question}
\cite{Amb:c:96}
Is it true that there is a constant $c>1$ such that
every interval $[x, y]\subseteq [\frac{1}{n}, \frac{1}{n-1}]$
with $y-x\geq (\frac{1}{c})^n$ contains at least one
point from the $\PFIN$-hierarchy?
\end{Question}

A similar question can be asked about $\FIN$.

\section{Other paradigms of inductive inference}
\label{lang}

Similar problems can be studied for other paradigms of inductive
inference (besides FIN). One of the most interesting open cases
is probabilistic language learning in the limit.
In language learning, the object to be learned is a 
recursively enumerable language $L$.
The standard presentation for the language is a {\em text.}

\begin{Definition}
\cite{Gol:j:67}
\begin{itemize}
\item[(a)]
A text $T$ for a language $L$ is an enumeration (in any order) of all words
in $L$.
\item[(b)]
A learning machine $M$ TxtEx-identifies a language $L$, if given any text $T$
for $L$ as an input, it outputs a sequence of grammars $g_1, g_2, \ldots$
such that $g_{i}=g_{i+1}=g_{i+2}=\ldots$ and $g_i$ recognizes 
$L$, for some $i\in\bbbn$.
\item[(c)]
A set $U$ of languages is called TxtEx-identifiable 
(identifiable in the limit) if 
there is a machine $M$ that TxtEx-identifies every $L\in U$.
TxtEx denotes the collection of all TxtEx-learnable sets.
\end{itemize}
\end{Definition}

Note that $M$ does not get any information
about the words not in $L$. This is the biggest difference
between inductive inference of functions and languages.
If $f(x)\neq y$, $M$ knows that after receiving $f(x)$.
If a word $x$ is not in $L$, $M$ never knows it because
this may be the case that $x\in L$ but it has not appeared in
the input yet.

The probability hierarchy for TxtEx has been
studied by Jain and Sharma\cite{Jai-Sha:j:96:team-lang}.
Below, we summarize their main results.
The first theorem concerns the probability at which
a probabilistic machine becomes stronger than 
a deterministic one. Similarly to PFIN or FIN, it
is 2/3. However, the similarities end once the next point in the
probability hierarchy is revealed. It is 5/8 (instead of 3/5). 
It remains open what are the next points below 5/8.

\begin{Theorem}
\cite{Jai-Sha:j:96:team-lang}
\begin{enumerate}
\item[(a)]
If $p>2/3$, then $\langle p\rangle\TXTEX=\TXTEX$.
\item[(b)]
$[2, 3]\TXTEX\neq\TXTEX$.
\item[(c)]
If $5/8<p\leq 2/3$, then $\langle p\rangle\TXTEX\neq [2, 3]\TXTEX$.
\item[(d)]
$[5, 8]\TXTEX\neq [2, 3]\TXTEX$.
\end{enumerate}
\end{Theorem}

The second theorem concerns relationships between teams of different
size at probability 1/2. Teams of different size may have
different learning power (similarly to FIN or PFIN). 
Also similarly to PFIN and FIN, all $[n, 2n]$-team sizes 
for odd $n$ were equivalent. However, for even $n$ results 
were no longer the same as for FIN or PFIN.

\begin{Theorem}
\cite{Jai-Sha:j:96:team-lang}
\begin{enumerate}
\item
$[1, 2]\TXTEX\neq [2, 4]\TXTEX$.
\item
$[1, 2]\TXTEX=[3, 6]\TXTEX=[5, 10]\TXTEX=\ldots$.
\item
For all $i\geq 0$, $[2^i, 2^{i+1}]\TXTEX\neq [2^{i+1}, 2^{i+2}]\TXTEX$.
\item
For all $k$, $[k, 2k]\TXTEX\neq \langle 1/2\rangle\TXTEX$.
\end{enumerate}
\end{Theorem}

In both cases, we see that there both similarities with FIN and
differences. We think that it can be very interesting to study this hierarchy.
However, before general results are proved, it may be necessary to
get a better knowledge of explicit probabilities and to accumulate
more proof techniques.

\section{Conclusions and related work}
\label{Related}

We surveyed the work in probabilistic inductive inference,
with an emphasis on recent work for FIN and PFIN.
For good surveys about earlier results,
see \cite{Ang-Smi:j:83,Osh-Sto-Wei:b:86:stl} for inductive inference
in general and \cite{Smi:c:94,Jai-Sha:c:95:teamgos} for
probabilistic inductive inference. 

The biggest challenge in the area remains obtaining general results about
the probability hierarchy for unrestricted $\FIN$.
In section \ref{FGen}, we mentioned several approaches 
to this problem. None of them has been completely successful
but there is a chance that these ideas can be extended,
giving more insight about unrestricted $\FIN$. 
There are other interesting problems about $\FIN$ that deserve 
studying as well (like $\FIN$ with oracles, section \ref{oracle}).

Besides $\FIN$, probability hierarchies for other learning models
can be studied. Probabilistic language learning\cite{Jai-Sha:j:96:team-lang} 
in the limit is the most interesting among those. 
Other recently studied models are probabilistic language learning
with monotonicity restrictions\cite{Mey:j:97} and probabilistic learning up 
to a small set of errors\cite{Vik:c:96}.

\bibliography{book}
\end{document}